\documentclass[10pt, a4paper]{article}
\usepackage[final]{lrec-coling2024}
\makeatletter
\def\@mb@citenamelist{cite,citep,citet,citealp,citealt,citepalias,citetalias}
\makeatother
\newcites{languageresource}{~}

\usepackage{graphicx}
\usepackage{tabularx}
\usepackage{soul}
\usepackage{placeins}

\usepackage{xcolor}
\usepackage{hyperref}
 \definecolor{darkblue}{rgb}{0, 0, 0.5}
  \hypersetup{colorlinks=true, citecolor=darkblue, linkcolor=darkblue, urlcolor=darkblue}

\usepackage{color}

\usepackage{amsmath}
\usepackage{amssymb}
\usepackage{amsfonts}
\usepackage[ruled, linesnumbered]{algorithm2e}
\usepackage{nicefrac}

\usepackage{color, soul}
\usepackage{lipsum}
\usepackage{multicol, multirow,makecell}
\usepackage{graphics}
\usepackage{graphicx}
\usepackage{booktabs}

\usepackage{comment}

\newcommand{\modelname}{Topics as Entity Clusters}
\newcommand\entity[1]{\textsc{#1}}

\newcommand{\pluseq}{\mathrel{+}=}
\renewcommand\subsubsection[1]{\textbf{#1.}}
\newcommand*{\tran}{^{\mkern-1.5mu\mathsf{T}}}

\title{Topics as Entity Clusters: Entity-based Topics from Large Language Models and Graph Neural Networks}

\name{Manuel V. Loureiro, Steven Derby, Tri Kurniawan Wijaya} 

\address{Huawei Ireland Research Centre\\  Dublin, Ireland\\ \{manuel.loureiro, tri.kurniawan.wijaya\}@huawei.com, steven.derby@huawei-partners.com}

\abstract{
Topic models aim to reveal latent structures within a corpus of text, typically through the use of term-frequency statistics over bag-of-words representations from documents. In recent years, conceptual entities --- interpretable, language-independent features linked to external knowledge resources --- have been used in place of word-level tokens, as words typically require extensive language processing with a minimal assurance of interpretability.
However, current literature is limited when it comes to exploring purely entity-driven neural topic modeling. For instance, despite the advantages of using entities for eliciting thematic structure, it is unclear whether current techniques are compatible with these sparsely organised, information-dense conceptual units.
In this work, we explore entity-based neural topic modeling and propose a novel topic clustering approach using bimodal vector representations of entities. Concretely, we extract these latent representations from large language models and graph neural networks trained on a knowledge base of symbolic relations, in order to derive the most salient aspects of these conceptual units.
Analysis of coherency metrics confirms that our approach is better suited to working with entities in comparison to state-of-the-art models, particularly when using graph-based embeddings trained on a knowledge base.
    \\ \newline \Keywords{neural topic modeling, entity clustering, knowledge base, large language model, graph neural network}
}

\begin{document}

\maketitleabstract

\section{Introduction}
\label{sec:introduction}
Following the seminal work of \citet{blei2003latent}, topic models have since become the \textit{de facto} method for extracting and elucidating prominent themes from corpora. Traditionally, the semantic content of a document is composed of document-term frequencies or latently through a mixture of distributions of topics, common with probabilistic generative models such as \emph{Latent Dirichlet Allocation} (LDA). Here, individual topics are represented by salient lexical constituents such as words that depict some subjects of the corpora \citep{blei2003latent, blei2006correlated, li2006pachinko, teh2006hierarchical, crain2012dimensionality}. In recent years, the field of \emph{Natural Language Processing} (NLP) has seen a trend toward continuous vector representations of words, which look to capture the paradigmatic relationship between concepts by learning distributional co-occurrence patterns in text. For example, large-scale language models such as \emph{BERT} \citep{devlin2018bert} produce robust contextualized representations that capture an array of linguistic phenomena and implicit real-world knowledge \citep{peters2018dissecting, tenney2019bert, tenney2018you, petroni2019language, rogers2020primer}, paving the way for so-called \emph{Neural Topic Modeling} \citep{sia2020tired, bianchi2020pre, grootendorst2022bertopic}. 

Despite their successes, it becomes evident that certain limitations emerge from conventional topic modeling due to the noisy and superfluous nature of word-level tokens. 
These methods rely heavily on text processing and data-driven techniques to uncover statistical patterns and infer relevant thematic structure within the corpus, resulting in topics with limited guarantees of expressiveness or interpretability.
An alternative approach, which has proven successful, involves extracting conceptual entities from the text and representing them with informative language-agnostic entity identifiers. These identifiers can be unambiguously linked to external lexical resources, encapsulating grounded real-world knowledge \citep{newman2006analyzing, chemudugunta2008modeling, andrzejewski2009incorporating, andrzejewski2011framework, kim2012etm, hu2013incorporating, allahyari2016discovering}. Entities, thus, provide rich, interpretable features that facilitate the elicitation of complex topical structure, particularly in multilingual settings, --- a limitation of current word-level topic models \citep{ni2009mining, boyd2009multilingual}.
However, despite these advantages, there is limited literature on the subject of purely entity-driven neural topic modeling; it is arguable whether conventional topic modeling approaches are suitable for adequately handling these sparse, information-rich data formats \citep{wang2006topic, zhao2010topic, hong2010study, quan2015short, li2016topic}. 

In this paper, we explore entities exclusively as topic features; conceptual entities are distinct, free-form human-derived concepts we represent using encyclopedic-based definitions and several key relational attributes, which offer a better alternative for topic modeling.
Here, we propose \emph{\modelname{}} (TEC)\footnote{Code available at \url{https://github.com/manuelvloureiro/topics-as-entity-clusters}}, a novel topic modeling algorithm that can discover meaningful and highly informative topics using vector representation clustering methods. 
We achieve this by using two sources to represent entities: (1) contextualized text representations constructed from entity definitions and (2) structured graph data extracted from a knowledge base that we use to train a graph neural network to learn node embeddings. 

Through the use of experimental procedure, we demonstrate that this combination of implicit contextualised information and explicit semantic
knowledge, in conjunction with our novel technique, are important components in overcoming similar data sparsity issues that are observed in other state-of-the-art topic models. Most impressively, we find that simpler graph-based embeddings trained on a knowledge graph of concepts demonstrate considerable improvements in comparison to the large language model embeddings, which we demonstrate across a range of coherency metrics.
\section{Literature Review}
\label{sec:literaturereview}

We are by no means the first to consider entity-based topic models. For instance, \citet{newman2006analyzing} proposed representing documents with salient entities obtained using \textit{Named Entity Recognition} (NER) instead of using the words directly, while entity-centric approaches have also been proposed which focus on word-entity distribution to generate topics \citep{kim2012etm, hu2013incorporating, allahyari2016discovering}. Others have attempted to capture the patterns among words, entities, and topics, either by expanding LDA \citep{blei2003latent} or more complex Bayesian topic models --- see \citet{alghamdi2015survey}, \citet{chauhan2021topic} and \citet{vayansky2020review} for a general overview.

\subsection{Word embeddings}
Researchers have also found success by capitalising on contemporary work in distributional semantics by integrating embedding lookup tables into their frameworks to represent words and documents.
For instance, \emph{lda2vec} \citep{moody2016mixing} combines embeddings with topic models by embedding word, document, and topic vectors into a common representation space.
Concurrently, \citet{van2016learning} introduces an unsupervised model that learns unidirectional mappings between latent vector representations of words and entities.
Using a shared embedding space for words and topics, \citet{dieng2020topic} instead presents the \emph{Embedded Topic Model} (ETM), which merges traditional topic models with the neural-based word embeddings of \citet{mikolov2013distributed}.

\subsection{Neural topic models}
In recent years, researchers have also looked to incorporate modern deep learning techniques
that utilize contextualized representations in contrast to more traditional static embeddings \citep{grootendorst2022bertopic, zhao2021topic}. \citet{srivastava2016neural} propose \emph{ProdLDA}, a neural variational inference method for LDA that explicitly approximates the Dirichlet prior. Other models, however, such as \emph{Neural Variational Document Model} (NVDM) \citep{miao2017discovering}, employ a multinomial factor model of documents that uses amortized variational inference to learn document representations.
\citet{bianchi2020pre} expand on ProdLDA presenting \emph{CombinedTM}, which improves the model with contextualized embeddings. 

\subsection{Knowledge extraction}
More related to our work, \citet{piccardi2021crosslingual}  --- leveraging the self-referencing nature of Wikipedia --- define a cross-lingual topic model in which documents are represented by extracted and densified bags-of-links. 
The adoption of large-scale knowledge base systems has recently gained popularity in NLP as a way to directly inject knowledge into the model \citep{gillick2019learning, sun2020ernie, liu2020k}, particularly when concerning conceptual entities \citep{andrzejewski2009incorporating, andrzejewski2011framework, yang2015efficient, terragni2020constraint, terragni2020concept}.

\subsection{Clustering}
Clustering techniques have also proved effective for topic modeling. For instance, \citet{sia2020tired} introduces clustering to generate coherent topic models from word embeddings, lowering complexity and producing better runtimes compared to traditional topic modeling approaches.
\citet{thompson2020topic} experiment with different pretrained contextualised embeddings and demonstrate that clustering contextualised representations at the token level is indistinguishable from a Gibbs sampling state for LDA.
These findings were also recently corroborated by \citet{zhang2022neural} who cluster sentence embeddings and extract top topic words using TF-IDF to produce more coherent and diverse topics than neural topic models.
\section{\modelname{}}
\begin{figure*}
    \centering
    \includegraphics[width=\textwidth]{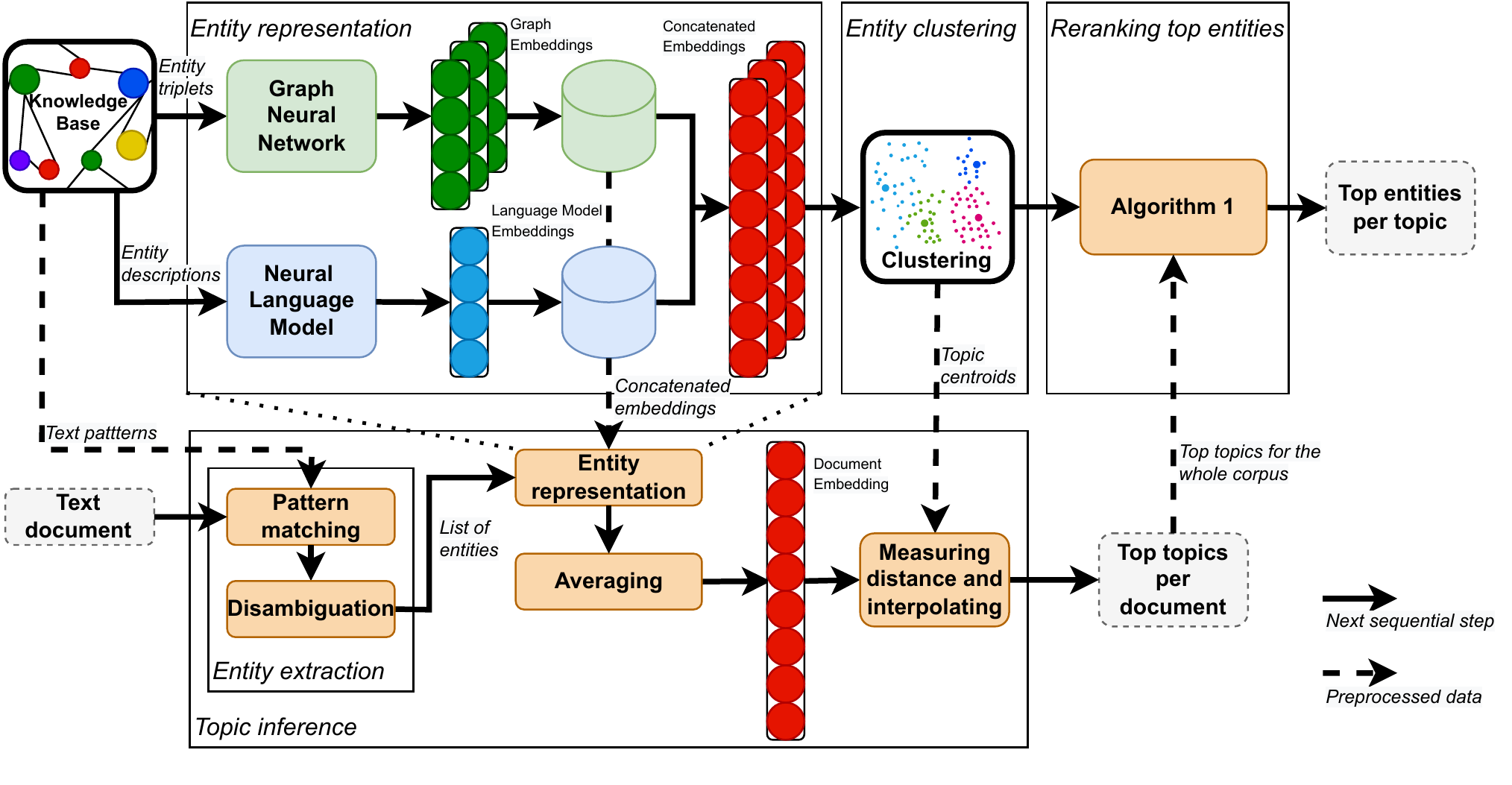}
    \vspace*{-5mm}
    \caption{Overview of \modelname{} (TEC). The top half illustrates the processing of entity embeddings, topic centroids and top entities per topic, while the bottom half inferencing the top topics per document.}
    \label{fig:diagram}
\end{figure*}

In this section, we describe the advantages and challanges associated with performing neural topic modeling with entities, as well as our proposed strategy. First, we begin by motivating our research before discussing our novel approach which can be viewed in Figure \ref{fig:diagram}.

\subsection{Background}
In the context of topic modeling, entities have proven highly useful to supplement or enrichen the data with expert symbolic data and provide factual grounding in the real world. In the literature, these approaches tend to either directly leverage entity data \citep{newman2006analyzing} or propose models that can effectively utilize prior knowledge in conjunction with word tokens \citep{kim2012etm, hu2013incorporating, yang2015efficient, terragni2020constraint, terragni2020concept}. In many ways, entities are well suited for this task, since topic models require the extrapolation of subjects through a small number of distinctly related concepts with variable overlapping themes. Furthermore, the use of words or phrase-level tokens, even in combination with external knowledge, are incompatible in multilingual settings, requiring techniques to link the languages together \citep{ni2009mining, boyd2009multilingual}, something that could be trivial with entity extraction and their language-agnostic representation through entity identifiers.

This approach is not without its caveats. A number of methods have been proposed to better accommodate entities in non-neural models, by either learning word-entity relationships or directly incorporating entity knowledge \citep{newman2006analyzing, chemudugunta2008modeling, andrzejewski2009incorporating, andrzejewski2011framework, allahyari2016discovering}. For one, documents may only contain a small number of entities, inhibiting the ability of the model to learn important document-term distributions with unsupervised techniques when taking an exclusively entity-based approach \citep{wang2006topic, zhao2010topic, hong2010study, quan2015short, li2016topic}. Several methods have been proposed to supplement the data in short text topic modeling by adding external information to the text \citep{hong2010study, weng2010twitter}, adding heuristic restrictions to the model \citep{nigam2000text} or altering the models to better handle sparse word co-occurrence patterns \citep{quan2015short, li2016topic, wang2021topic}. However, there is limited work pertaining to purely entity-driven neural topic models, which tend to either incorporate LDA into their approach --- which struggles in sparse data settings ---  or rely on term-frequencies for topic inference \citep{srivastava2016neural, miao2017discovering, bianchi2020pre, zhao2021topic, grootendorst2022bertopic}. Furthermore, measures used to analyse topic quality heavily rely on automated techniques that directly compare \emph{Pointwise Mutual Information} (PMI) with the resulting topic structure, which also depends on co-occurrence statistics. As such, it is important to consider the results of these metrics when performing our analysis \footnote{We also note that there are questions as to whether these PMI measures correlate with human intuition regardless \citep{hoyle2021broken, doogan2021twaddle}}.

\subsection{Entity representation}
\label{sec:entity_representation}

Despite the limited research on the subject, pretrained neural language models should be effective at supplementing topic models in minimal data setting due to their language capabilities and their rich understanding of real-world knowledge \citep{petroni2019language}. However, an approach for generating vector representations for producing the best topics is less clear, particularly when working with sparsely distributed and expressive entity-document features. Broadly, we can consider two methods for constructing entity representations effective for this data structure \citep{wang2017short}: \textit{implicit knowledge} from a pre-trained large language model and \textit{explicit knowledge} extracted directly from a knowledge graph.

\subsubsection{Implicit knowledge}
\label{sec:language-model-embeddings}
Throughout the field of natural language process, language models have been used to construct document representations with great success to express salient knowledge obtained implicitly through a considerable amount of unsupervised learning \citep{petroni2019language}.
However, it would be suboptimal to simply compute documents embeddings based on the bag-of-entities approach, as the response elicited from unsupervised models are highly contextualized in nature \citep{ethayarajh2019contextual}. We can overcome this challenge by instead extracting descriptions of the entities from an encyclopedic corpus --- another benefit of using human-curated conceptual entities.

\subsubsection{Explicit knowledge}
An advantage of using entities from lexical resources such as a knowledge base is that they provide a systematic framework for organizing and describing human-annotated relationships between concepts. Indeed, the entities confined within these semantic networks exhibit complex relational structures that provide meaningful information about their content, provided in the form of a directed graph. For instance, the triplet
\entity{<Petra, Culture, Nabataean kingdom>}
contains intricate encyclopedic knowledge about the city of Petra that can be difficult to learn with less specialised corpora. Language models may fail to adequately capture this relationship due to the abstract notion of the concept. In addition, explicit knowledge relations afford us supplementary information which is important for sparse bag-of-entities representations.

\subsubsection{Combining approaches}
\label{sec:approach}
Since both methods provide their own distinct advantages to modeling entities, we experiment with combining both sources of knowledge in order to accurately and effectively represent conceptual entities.
For some normalized language and graph embeddings $\hat{E}_{LM} \in \mathbb{R}^{d_{LM}}$ and $\hat{E}_{G} \in \mathbb{R}^{d_{G}}$, respectively, we weight their contributions using the following concatenation function,
\begin{equation}
    \hat{E} = \left[\sqrt{\frac{1}{1+\alpha}}\cdot \hat{E}_{LM}\tran, \sqrt{\frac{\alpha}{1+\alpha}}\cdot \hat{E}_{G}\tran \right]\tran \label{eq:concat}
\end{equation}
where $\alpha \in \mathbb{R}$ is the scalar rusatio of embedding weights and $\hat{E} \in \mathbb{R}^{d_{LM} + d_{G}}$ is our final embedding used in entity clustering.
We take the square root to guarantee that the final embedding is normalized similarly to the input embeddings.

\subsection{Entity clustering}
\label{sec:entity-clustering}
Independent of the specific method, in this work we represent entities in an embedding space and use clustering to construct centroids which we interpret as topic centroids. As such, we model topics based on the innate structure that emerges from the shared embedding space with these entities, similar to other methods in the literature. To this effect, we apply K-Means to the set of entities contained in a corpus, using the implementation available in FAISS \citep{johnson2019billion}.
We adopt a two-stage approach to extract entities, which allows us to represent text as a language-agnostic collection of entity identifiers arranged in order of appearance. We first extract candidate entities by finding language-specific text patterns in the original text, followed by a disambiguation process. We include further in-depth detail in Appendix \ref{app:entity-extraction}.

\subsection{Topic inference}
\label{sec:topic_inference}
Topic inference requires that the representation of documents be in the same embedding space as entities and topic centroids. To accomplish this, we extract entities as described in Section \ref{sec:entity-clustering}. We then obtain the document representation by calculating the weighted average of those entity embeddings, weighted by term frequency. That is, for each term $t$ in a document $d$, we generate the document embedding as,

\begin{equation}
    E_d = \dfrac{1}{|d|}\sum_{t \in d} tf_{t, d} \cdot E_{t}
\end{equation}

where $tf$ is our term-frequency matrix and $E_{t}$ is the embedding of the term generated as described in section \ref{sec:approach}.
With $K$ representing the number of topics, we can now measure the Euclidean distances $\textbf{d}=\left[d_1, d_2, ..., d_K\right]\tran$ of the document to the topic centroids. Documents are assumed to contain a share of all topics. We infer the topic weight contribution $\textbf{w}=\left[w_1, w_2, ..., w_K\right]\tran$ to the document using the inverse distance squared weighted interpolation. If we consider the embedding of a document as an interpolation of topic centroids, squaring the distances yields more weight to the closest topic centroids, which helps when working with irregularly-spaced data \citep{shepard1968two}:
\begin{equation}
    w_i = \frac{d_i^{-2}}{\sum_{j=1}^K d_j^{-2}} \quad , \forall\, i \in\left\lbrace1 , ...\, , K\right\rbrace .
\end{equation}

\subsection{Reranking top entities}
A list of highly descriptive entities, weighted by their importance, can be used to express the theme of a topic. However, the closest entities to topic centroids are not necessarily the most descriptive, due to the fact that this naive strategy disregards import entity co-occurrence information within the corpus. To help alleviate issues associated with data sparsity we propose a novel inference-based method to rerank top entities, which assigns the entity frequency of a document to the top topic centroid, as measured by $\textbf{w}$ (see Algorithm \ref{alg:top_words}).

We start by assigning entities to topics based on their distances weighted by a small value, $\epsilon$ (Lines \ref{alg:line:closest_start}-\ref{alg:line:closest_end}).
We follow by inferring the top topic for each document and updating the top entities in that topic using the document entity frequency. The update is proportional to the inference score, $\max\left(\textbf{w}\right)$, as it represents the degree of confidence in the inference (Lines \ref{alg:line:inference_start}-\ref{alg:line:inference_end}). To further improve topic diversity in this sparse setting, we only update the top topic.
Lastly, we calculate the relative frequencies to obtain the top entities per topic (Lines \ref{alg:line:relative_start}-\ref{alg:line:relative_end}).

\begin{algorithm}
    \small
    \KwIn{Number of topics $K$, number of top entities per topic $N$, small initialization weight  $\epsilon$, documents $\mathit{Docs}$, all entity identifiers in the corpus $\mathit{entities}$, entity embeddings $\hat{E}$}
    \KwOut{Lists of top entities per topic $\mathit{topEntities}$, each element is a list of pairs $(\mathit{entityId}, \mathit{frequency})$}
    \For{$\mathit{topicId} \in \left\lbrace1, ..., K\right\rbrace$ \label{alg:line:closest_start}}
    {
        $\mathit{topEntities}[\mathit{topicId}] \gets \operatorname{ClosestEntities}(\mathit{topicId}, \hat{E}, N, \epsilon)$

    } \label{alg:line:closest_end}
    \For{$\mathit{doc} \in \mathit{Docs}$ \label{alg:line:inference_start}}{
        $\textbf{w},\, \mathit{entityFrequency}  \gets \operatorname{TopicInference}(\mathit{doc})$ 
        $\mathit{topTopic} \gets \operatorname{argmax(\textbf{w})}$\\
        \For{$\mathit{entityId} \in \mathit{entities}$}{
            $\mathit{topEntities}\left[\mathit{topTopic}\right]\left[\mathit{entityId}\right]
                \pluseq\max\left(\textbf{w}\right) \cdot \mathit{entityFrequency}[\mathit{entityId}]$
        }
    } \label{alg:line:inference_end}
    \For{$\mathit{topicId} \in \left[1, ..., K\right]$ \label{alg:line:relative_start}}{
        $\mathit{topEntities}[\mathit{topicId}] \gets \operatorname{RelativeFrequency}(\mathit{topEntities}[\mathit{topicId}])$
    } \label{alg:line:relative_end}
    \caption{Reranking top entities}
    \label{alg:top_words}
\end{algorithm}
\begin{table}
    \centering
    \caption{Statistics of the corpora.}
    \label{table:corpora_statistics}
    \resizebox{\columnwidth}{!}{%
        \begin{tabular}{lrrr}
            \toprule
            \textbf{Corpus} & Vocabulary & Documents & \multicolumn{1}{c}{\makecell{Avg. Entities \\ per Document}} \\
            \midrule
            WIKIPEDIA       & 359,507    & 359,507   & 44.62                                      \\
            CC-NEWS         & 94,936     & 412,731   & 13.97                                      \\
            MLSUM           & 89,383     & 661,422   & 11.71                                      \\
            \bottomrule
        \end{tabular}
    }
\end{table}

\label{exp}
\section{Experiments}
We study the performance of TEC and qualitatively compare it to other state-of-the-art topic models using a set of corpora preprocessed into lists of entity identifiers. By contrasting the top entities and measuring results across several metrics, we can infer the quality of each topic model. As it does not pertain to our work, we do not directly compare models trained on entities and word-level tokens, although we do provide an in-depth discussion in Appendix \ref{app:words_ents} on why the comparison between word and entity-topic models is challenging, as well as some experimental analysis. Furthermore, since representations are constructed from language-agnostic entity-identifiers, we cannot partition the results by the source languages as our approach is invariant in this setting.

\subsection{Entity selection and embedding}
We build the entity extractor using Wikidata\footnote{ \href{https://dumps.wikimedia.org/wikidatawiki/entities/latest-all.json.bz2}{Wikidata JSON dump}}
as the source of our knowledge base and the English version of Wikipedia\footnote{Collected with  \href{https://www.crummy.com/software/BeautifulSoup/bs4/doc/}{Beautiful Soup}}
as the encyclopedic corpus. Wikidata entities contain a reference to the corresponding Wikipedia article.  Wikidata currently has more than 97 million entities, most of which would be a long tail of entities in a topic model therefore we restrict the entity extractor to only include the top one million entities, as ranked by QRank\footnote{\href{https://qrank.wmcloud.org/}{QRank}}
-- a public domain project that ranks page views across Wikimedia projects.
Out of these entities, we select those matching at least one predicate-object pair from lists of preselected objects for predicates \entity{Instance Of}, \entity{Subclass Of}, and \entity{Facet Of}.

Specifically for the graph neural network, and considering the previously selected entities, we then incorporate all predicates that reference objects as entities existing in the knowledge base. This strategy not only enhances the graph's density but also contributes to more robust distinctions between similar entities, resulting in richer embedding representations. For instance, an entity like \entity{<Jack Dorsey, Occupation, Entrepreneur>} is included since we have \entity{<Entrepreneur, Instance Of, Profession>} in our knowledge base. Conversely, entries like \entity{<Jack Dorsey, Eye Color, Light Brown>} are excluded, as they do not match the initial filtering criteria.

Entities are then encoded both using a large language model and a graph neural network. Not withstanding that other models could be used, we use SBERT\footnote{We use \href{https://huggingface.co/sentence-transformers/paraphrase-multilingual-mpnet-base-v2}{paraphrase-multilingual-mpnet-base-v2}.} and \emph{node2vec} \citep{grover2016node2vec} respectively.

\subsection{Corpora}

We evaluate all models on various corpora: \emph{Wikipedia},  \emph{CC-News}\footnote{\emph{CC-News} available at \href{https://huggingface.co/datasets/cc_news}{Hugging Face}.}, and \emph{MLSUM} \citep{scialom2020mlsum}; Table \ref{table:corpora_statistics} contains a statistics summary. The \emph{Wikipedia} corpus consists of a sample of preprocessed documents, each matching an entity in the vocabulary. \emph{CC-News} consists of monolingual news articles written in English.
\emph{MLSUM} on the other hand is a collection of news articles written in multiple languages: German, Spanish, French, Russian, and Turkish. Topic modeling across multiple languages faces challenges due to the language bifurcation problem, which refers to the emergence of thematic structures that are segmented by their source language \citep{boyd2009multilingual}. Despite this, the language-agnostic nature of entities offers a unique advantage, unaffected by such issues, compared to word tokens. For instance, the entity \entity{Earth} is given by a tag \textbf{(Q2)} rather than the singular lexical unit "Earth" in English, "Terre" in French or "Erde" in German.

We preprocess the documents according to Section \ref{sec:entity-clustering}. The language-specific components for documents in English, German, Spanish and French are \textit{spaCy} lemmatizers \citep{spacy2}, for documents in Russian we use \textit{pymorphy2} \citep{korobov2015morphological}, and for documents in Turkish we use \textit{zeyrek}\footnote{\emph{zeyrek} available on \url{https://github.com/obulat/zeyrek}.}.

\begin{table*}[!ht]
    \scriptsize
    \centering
    \setlength{\tabcolsep}{2pt}
    \begin{tabular}{l|rrr|rrr|rrr}
        \toprule
                                               & \multicolumn{3}{c|}{$\mathbf{WIKIPEDIA}$} & \multicolumn{3}{c|}{$\mathbf{CC-NEWS}$} & \multicolumn{3}{c}{$\mathbf{MLSUM}$}                                                                                                                                                                                                                            \\
        \toprule
        \textbf{Languages}                     & \multicolumn{3}{c|}{en}                   & \multicolumn{3}{c|}{en}                 & \multicolumn{3}{c}{fr, de, es, ru, tr}                                                                                                                                                                                                                          \\ \midrule
        \textbf{Metric}                        & \multicolumn{1}{c}{$\mathbf{TC}$}         & \multicolumn{1}{c}{$\mathbf{TD}$}       & \multicolumn{1}{c|}{$\mathbf{TQ}$}     & \multicolumn{1}{c}{$\mathbf{TC}$} & \multicolumn{1}{c}{$\mathbf{TD}$} & \multicolumn{1}{c|}{$\mathbf{TQ}$} & \multicolumn{1}{c}{$\mathbf{TC}$} & \multicolumn{1}{c}{$\mathbf{TD}$} & \multicolumn{1}{c}{$\mathbf{TQ}$} \\
        \midrule \specialrule{.1em}{.05em}{.05em}
        \multicolumn{4}{l}{\textbf{Number of Topics} $\mathbf{\times 100}$}                                                                                                                                                                                                                                                                                                                            \\
        \midrule
        LDA (Entities)                         & $-0.05$ \tiny{$(0.01)$}                   & $\textbf{0.98}$ \tiny{$(0.00)$}         & $-0.05$ \tiny{$(0.01)$ }               & $-0.13$ \tiny{$(0.01)$}           & $\textbf{0.97}$ \tiny{$(0.00)$}   & $-0.13$ \tiny{$(0.01)$ }           & $-0.02$ \tiny{$(0.01)$}           & $\textbf{0.96}$ \tiny{$(0.00)$}   & $-0.02$ \tiny{$(0.01)$ }          \\
        NVDM-GSM                               & $0.06$ \tiny{$(0.02)$}                    & $0.87$ \tiny{$(0.02)$}                  & $0.05$ \tiny{$(0.02)$ }                & $-0.03$ \tiny{$(0.02)$}           & $0.61$ \tiny{$(0.14)$}            & $-0.02$ \tiny{$(0.01)$ }           & $0.08$ \tiny{$(0.01)$}            & $0.59$ \tiny{$(0.09)$}            & $0.04$ \tiny{$(0.01)$ }           \\
        ProdLDA                                & $-0.16$ \tiny{$(0.03)$}                   & $0.62$ \tiny{$(0.16)$}                  & $-0.10$ \tiny{$(0.03)$ }               & $-0.30$ \tiny{$(0.01)$}           & $0.23$ \tiny{$(0.01)$}            & $-0.07$ \tiny{$(0.00)$ }           & $-0.21$ \tiny{$(0.02)$}           & $0.36$ \tiny{$(0.04)$}            & $-0.08$ \tiny{$(0.01)$ }          \\
        CombinedTM                             & $-0.10$ \tiny{$(0.02)$}                   & $0.22$ \tiny{$(0.03)$}                  & $-0.02$ \tiny{$(0.00)$ }               & $-0.32$ \tiny{$(0.01)$}           & $0.37$ \tiny{$(0.21)$}            & $-0.12$ \tiny{$(0.07)$ }           & $-0.22$ \tiny{$(0.01)$}           & $0.37$ \tiny{$(0.11)$}            & $-0.08$ \tiny{$(0.02)$ }          \\
        WikiPDA                                & $0.08$ \tiny{$(0.01)$}                    & $0.73$ \tiny{$(0.01)$}                  & $0.06$ \tiny{$(0.00)$ }                & \multicolumn{1}{c}{---}           & \multicolumn{1}{c}{---}           & \multicolumn{1}{c|}{---}           & \multicolumn{1}{c}{---}           & \multicolumn{1}{c}{---}           & \multicolumn{1}{c}{---}           \\
        TEC $E_{LM} \left(\alpha=0\right)$     & $0.18$ \tiny{$(0.01)$}                    & $0.95$ \tiny{$(0.00)$}                  & $0.17$ \tiny{$(0.01)$ }                & $0.11$ \tiny{$(0.01)$}            & $0.79$ \tiny{$(0.01)$}            & $0.08$ \tiny{$(0.01)$ }            & $0.16$ \tiny{$(0.01)$}            & $0.79$ \tiny{$(0.01)$}            & $0.13$ \tiny{$(0.01)$ }           \\
        TEC $\alpha=\nicefrac{1}{2}$           & $0.21$ \tiny{$(0.01)$}                    & $0.95$ \tiny{$(0.01)$}                  & $0.20$ \tiny{$(0.01)$ }                & $0.17$ \tiny{$(0.01)$}            & $0.81$ \tiny{$(0.01)$}            & $0.14$ \tiny{$(0.01)$ }            & $\textbf{0.24}$ \tiny{$(0.01)$}   & $0.82$ \tiny{$(0.01)$}            & $0.19$ \tiny{$(0.01)$ }           \\
        TEC $\alpha=1$                         & $0.21$ \tiny{$(0.01)$}                    & $0.96$ \tiny{$(0.01)$}                  & $0.21$ \tiny{$(0.01)$ }                & $0.19$ \tiny{$(0.02)$}            & $0.82$ \tiny{$(0.01)$}            & $0.15$ \tiny{$(0.02)$ }            & $\textbf{0.24}$ \tiny{$(0.01)$}   & $0.82$ \tiny{$(0.01)$}            & $0.19$ \tiny{$(0.01)$ }           \\
        TEC $\alpha=2$                         & $0.22$ \tiny{$(0.01)$}                    & $0.97$ \tiny{$(0.00)$}                  & $0.22$ \tiny{$(0.01)$ }                & $0.19$ \tiny{$(0.01)$}            & $0.83$ \tiny{$(0.01)$}            & $\textbf{0.16}$ \tiny{$(0.01)$ }   & $\textbf{0.24}$ \tiny{$(0.01)$}   & $0.82$ \tiny{$(0.01)$}            & $\textbf{0.20}$ \tiny{$(0.01)$ }  \\
        TEC $E_{G} \left(\alpha=\infty\right)$ & $\textbf{0.24}$ \tiny{$(0.01)$}           & $0.97$ \tiny{$(0.00)$}                  & $\textbf{0.23}$ \tiny{$(0.01)$ }       & $\textbf{0.20}$ \tiny{$(0.02)$}   & $0.83$ \tiny{$(0.01)$}            & $\textbf{0.16}$ \tiny{$(0.02)$ }   & $\textbf{0.24}$ \tiny{$(0.01)$}   & $0.83$ \tiny{$(0.01)$}            & $\textbf{0.20}$ \tiny{$(0.01)$ }  \\
        \midrule \specialrule{.1em}{.05em}{.05em}
        \multicolumn{4}{l}{\textbf{Number of Topics} $\mathbf{\times 300}$}                                                                                                                                                                                                                                                                                                                            \\
        \midrule
        LDA (Entities)                         & $0.09$ \tiny{$(0.01)$}                    & $\textbf{0.97}$ \tiny{$(0.00)$}         & $0.09$ \tiny{$(0.01)$ }                & $-0.07$ \tiny{$(0.01)$}           & $\textbf{0.91}$ \tiny{$(0.00)$}   & $-0.07$ \tiny{$(0.01)$ }           & $0.03$ \tiny{$(0.01)$}            & $\textbf{0.88}$ \tiny{$(0.00)$}   & $0.02$ \tiny{$(0.01)$ }           \\
        NVDM-GSM                               & $0.06$ \tiny{$(0.02)$}                    & $0.69$ \tiny{$(0.03)$}                  & $0.04$ \tiny{$(0.02)$}                 & $0.04$ \tiny{$(0.01)$}            & $0.52$ \tiny{$(0.05)$}            & $0.02$ \tiny{$(0.01)$}             & $0.13$ \tiny{$(0.01)$}            & $0.42$ \tiny{$(0.04)$}            & $0.06$ \tiny{$(0.01)$ }           \\
        ProdLDA                                & $-0.14$ \tiny{$(0.02)$}                   & $0.44$ \tiny{$(0.17)$}                  & $-0.06$ \tiny{$(0.03)$ }               & $-0.21$ \tiny{$(0.01)$}           & $0.16$ \tiny{$(0.01)$}            & $-0.03$ \tiny{$(0.00)$ }           & $-0.16$ \tiny{$(0.01)$}           & $0.17$ \tiny{$(0.01)$}            & $-0.03$ \tiny{$(0.00)$ }          \\
        CombinedTM                             & $-0.13$ \tiny{$(0.03)$}                   & $0.15$ \tiny{$(0.03)$}                  & $-0.02$ \tiny{$(0.00)$ }               & $-0.32$ \tiny{$(0.01)$}           & $0.41$ \tiny{$(0.17)$}            & $-0.13$ \tiny{$(0.05)$ }           & $-0.23$ \tiny{$(0.01)$}           & $0.17$ \tiny{$(0.07)$}            & $-0.04$ \tiny{$(0.02)$ }          \\
        WikiPDA                                & $0.06$ \tiny{$(0.01)$}                    & $0.84$ \tiny{$(0.01)$}                  & $0.05$ \tiny{$(0.00)$ }                & \multicolumn{1}{c}{---}           & \multicolumn{1}{c}{---}           & \multicolumn{1}{c|}{---}           & \multicolumn{1}{c}{---}           & \multicolumn{1}{c}{---}           & \multicolumn{1}{c}{---}           \\
        TEC $E_{LM} \left(\alpha=0\right)$     & $0.25$ \tiny{$(0.01)$}                    & $0.95$ \tiny{$(0.00)$}                  & $0.24$ \tiny{$(0.01)$ }                & $0.11$ \tiny{$(0.01)$}            & $0.72$ \tiny{$(0.00)$}            & $0.08$ \tiny{$(0.01)$ }            & $0.14$ \tiny{$(0.01)$}            & $0.74$ \tiny{$(0.01)$}            & $0.10$ \tiny{$(0.01)$ }           \\
        TEC $\alpha=\nicefrac{1}{2}$           & $0.29$ \tiny{$(0.01)$}                    & $0.95$ \tiny{$(0.00)$}                  & $0.28$ \tiny{$(0.01)$ }                & $\textbf{0.18}$ \tiny{$(0.01)$}   & $0.75$ \tiny{$(0.01)$}            & $0.13$ \tiny{$(0.01)$ }            & $0.22$ \tiny{$(0.01)$}            & $0.76$ \tiny{$(0.00)$}            & $0.16$ \tiny{$(0.01)$ }           \\
        TEC $\alpha=1$                         & $0.30$ \tiny{$(0.01)$}                    & $0.96$ \tiny{$(0.00)$}                  & $0.29$ \tiny{$(0.01)$ }                & $\textbf{0.18}$ \tiny{$(0.01)$}   & $0.75$ \tiny{$(0.00)$}            & $\textbf{0.14}$ \tiny{$(0.01)$ }   & $0.22$ \tiny{$(0.01)$}            & $0.76$ \tiny{$(0.01)$}            & $\textbf{0.17}$ \tiny{$(0.01)$ }  \\
        TEC $\alpha=2$                         & $\textbf{0.31}$ \tiny{$(0.01)$}           & $0.96$ \tiny{$(0.00)$}                  & $\textbf{0.30}$ \tiny{$(0.01)$ }       & $\textbf{0.18}$ \tiny{$(0.01)$}   & $0.76$ \tiny{$(0.01)$}            & $\textbf{0.14}$ \tiny{$(0.00)$ }   & $0.22$ \tiny{$(0.01)$}            & $0.76$ \tiny{$(0.01)$}            & $\textbf{0.17}$ \tiny{$(0.01)$ }  \\
        TEC $E_{G} \left(\alpha=\infty\right)$ & $\textbf{0.31}$ \tiny{$(0.01)$}           & $0.96$ \tiny{$(0.00)$}                  & $\textbf{0.30}$ \tiny{$(0.01)$}        & $\textbf{0.18}$ \tiny{$(0.01)$}   & $0.76$ \tiny{$(0.01)$}            & $\textbf{0.14}$ \tiny{$(0.01)$}    & $\textbf{0.23}$ \tiny{$(0.01)$}   & $0.76$ \tiny{$(0.01)$}            & $\textbf{0.17}$ \tiny{$(0.00)$}   \\ \bottomrule
    \end{tabular}
    \caption{Results on all corpora and topic models. We record the results on $\mathbf{TC:}$ (Topic Coherence), based on the normalized pointwise mutual information of top $N=10$ entities in a topic, $\mathbf{TD}:$ (Topic Diversity) the ratio of unique entities to total entities and $\mathbf{TQ:}$ (Topic Quality) Topic Diversity $\times$ $\mathbf{TC}$. The results are reported as the averages of 10 experimental runs and include the margin of error related to 95\% confidence intervals.
        \label{tab:coherence}}
\end{table*}

\subsection{Models}
Although not strictly a neural topic model, we begin by comparing our approach with LDA \citep{blei2003latent} due to its pervasiveness in the literature. Specifically, we use the \emph{Mallet} implementation of LDA \citep{mccallum2002mallet}. On top of that, we compare using other state-of-the-art neural topic models from the literature.

\subsubsection{NVDM-GSM} \emph{Neural Variational Document Model} (NVDM) is a neural network-based topic model that discovers topics through variational inference training, proposing a number of ways to construct topic distributions, such as a \emph{Gaussian Softmax} (GSM) function  \citep{miao2017discovering}.

\subsubsection{ProdLDA} Similar to NVDM-GSM, this model is an autoencoder trained to reconstruct the input embeddings with variational inference-based training \citep{srivastava2016neural}.

\subsubsection{CombinedTM} This model is a direct extension to ProdLDA that includes pre-trained contextualized embeddings from a pretrained large language model \citep{bianchi2020pre}. It extracts contextual vectors for documents using \emph{SBERT}.

\subsubsection{WikiPDA} We also consider the Wikipedia-based Polyglot Dirichlet Allocation model, an LDA model trained on entities extracted from Wikipedia  \cite{piccardi2021crosslingual}. Although it is not a neural approach it is state-of-the-art for working with topic modeling on Wikipedia data, which has its own preprocessing method for further comparison.

\subsection{Metrics}

We want to evaluate how well models capture the meaning of topics using topic coherence ($\mathbf{TC}$), to estimate the relationship between top entities of a topic \citep{roder2015exploring}. We follow the implementation of \emph{gensim} \citep{rehurek2011gensim},

\begin{equation}
    \operatorname{TC} = \frac{1}{T} \sum_{t \in \left\lbrace 1..T\right\rbrace } \left[\frac{2}{ N ( N - 1)} \sum_{\substack{i \in \left\lbrace 1..N\right\rbrace \\ j \in \left\lbrace 1..i-1\right\rbrace }} \operatorname{NPMI}_{tij} \right]. \label{eq:coherence_per_topic}
\end{equation}

Topic coherence is defined over the most relevant entities for all topics $t \in \left\lbrace1..T\right\rbrace$, where $T$ is the number of topics, $N$ is the number of top entities per topic (we use $N=10$), and $\operatorname{NPMI}_{tij}$ is the normalized pointwise mutual information between entities $e_{t(i)}$ and $e_{t(j)}$, with $e_{t(i)}$ representing top entity of topic $t$ with index $i$. Normalized pointwise mutual information is given by:

\begin{equation}
    \operatorname{NPMI}_{tij} = \frac{ \log \frac{P(e_{t(i)}, \, \, e_{t(j)})}{P(e_{t(i)}) \,  \, P(e_{t(j)})}}{-\log\left(P(e_{t(i)},\, \,e_{t(j)})\right)}. \label{eq:npmi}
\end{equation}

We use normalized pointwise mutual information as it has been shown to correlate well with human judgments of topic coherence \citep{bouma2009normalized}.

Topic diversity ($\mathbf{TD}$) is the ratio between the number of unique entities and the total number of entities, considering the top 25 entities per topic \citep{dieng2020topic}.
Topic quality ($\mathbf{TQ}$) is the product of topic coherence and topic diversity.

\subsection{Implicit vs Explicit knowledge}
To further supplement our findings, we also perform experiments on TEC to determine how the influence of different types of knowledge affects its performance, in this case, the implicit or explicit knowledge from the language model and graph neural network trained on a knowledge base respectively. To achieve this, we train several models with varying values of $\alpha$, a parameter which effectively controls the impact of a particular source of information as described in Equation \ref{eq:concat}. As $\alpha \rightarrow \infty$, the influence of the language model representations on TEC will approach zero, leaving TEC to rely exclusively on the graph neural network representations. Conversely, setting $\alpha=0$ will ensure that TEC relies solely on the language model representations, excluding information from the graph neural network. In this work, we compare the default TEC model, which shares equal influence from both sources of information ($\alpha=1$), with models trained using varying degrees of information from each of these channels.

\subsection{Experiments specifications}

For each combination of model, corpus, and the number of topics, i.e., 100 and 300, we compute metrics over 10 runs and present both the averages and 95\% confidence interval margin of error in Table \ref{tab:coherence}.
We use sequential seeds for the sake of reproducibility.
We use implementation defaults for all models, with the exceptions of NVDM-GSM, where we run 100 epochs, and ProdLDA and CombinedTM, that we run each for 250 epochs.
We report metrics for the epoch with higher $\mathbf{TC}$.

We run the experiments in a shared Linux machine with 72 CPU cores, 256GB RAM and a Tesla V100-SXM2-16GB GPU.

\begin{table*}[!ht]
    \scriptsize
    \centering
    \begin{tabular}{l|l}
        \toprule
        \textbf{Model}                         & \textbf{Sample Topic}                                                                                                                                                                                                                                                                                                                                                                                                                                                                                                                                                                                                                           \\ \midrule \specialrule{.1em}{.05em}{.05em}
        LDA                                    & \multicolumn{1}{p{12cm}}{\scriptsize{\textbf{United Nations} (Q1065)\, |}\, \scriptsize{\textbf{Teenage Mutant Ninja Turtles} (Q12296099)\, |}\, \scriptsize{\textbf{Miles Davis} (Q93341)\, |}\, \scriptsize{\textbf{Star Trek} (Q1092)\, |}\, \scriptsize{\textbf{United Nations Security Council} (Q37470)\, |}\, \scriptsize{\textbf{United Nations Relief and Works Agency for Palestine Refugees in the Near East} (Q846656)\, |}\, \scriptsize{\textbf{public health} (Q189603)\, |}\, \scriptsize{\textbf{Dizzy Gillespie} (Q49575)\, |}\, \scriptsize{\textbf{Greenpeace} (Q81307)\, |}\, \scriptsize{\textbf{John Coltrane} (Q7346)}} \\
        NVDM-GSM                               & \multicolumn{1}{p{12cm}}{\scriptsize{\textbf{bitcoin} (Q131723)\, |}\, \scriptsize{\textbf{Apple Inc.} (Q312)\, |}\, \scriptsize{\textbf{Halloween {[}film franchise{]}} (Q1364022)\, |}\, \scriptsize{\textbf{Fisker Inc. {[}automaker{]}} (Q1420893)\, |}\, \scriptsize{\textbf{IBM} (Q37156)\, |}\, \scriptsize{\textbf{Michael Myers} (Q1426891)\, |}\, \scriptsize{\textbf{Yakuza {[}video game series{]}} (Q2594935)\, |}\, \scriptsize{\textbf{Facebook} (Q355)\, |}\, \scriptsize{\textbf{cryptocurrency} (Q13479982)\, |}\, \scriptsize{\textbf{Vancouver} (Q234053)}}                                                                 \\
        ProdLDA                                & \multicolumn{1}{p{12cm}}{\scriptsize{\textbf{Paul McCartney} (Q2599)\, |}\, \scriptsize{\textbf{Maxim Gorky} (Q12706)\, |}\, \scriptsize{\textbf{Lucy-Jo Hudson} (Q1394969)\, |}\, \scriptsize{\textbf{Bob Dylan}(Q392)\, |}\, \scriptsize{\textbf{sport utility vehicle} (Q192152)\, |}\, \scriptsize{\textbf{FIFA World Cup} (Q19317)\, |}\, \scriptsize{\textbf{sedan} (Q190578)\, |}\, \scriptsize{\textbf{American football} (Q41323)\, |}\, \scriptsize{\textbf{concept car} (Q850270)\, |}\, \scriptsize{\textbf{racing automobile} (Q673687)}}                                                                                          \\
        CombinedTM                             & \multicolumn{1}{p{12cm}}{\scriptsize{\textbf{vocalist} (Q2643890)\, |}\, \scriptsize{\textbf{United States of America} (Q30)\, |}\, \scriptsize{\textbf{music interpreter} (Q3153559)\, |}\, \scriptsize{\textbf{England} (Q21)\, |}\, \scriptsize{\textbf{Ryuichi Sakamoto} (Q345494)\, |}\, \scriptsize{\textbf{human rights} (Q8458)\, |}\, \scriptsize{\textbf{David Tennant} (Q214601)\, |}\, \scriptsize{\textbf{Harry Potter} (Q76164749)\, |}\, \scriptsize{\textbf{Comedian} (Q2591461)\, |}\, \scriptsize{\textbf{Aoni Production} (Q1359479)}}                                                                                       \\
        WikiPDA                                & \multicolumn{1}{p{12cm}}{\scriptsize{\textbf{a cappella} (Q185298)\, |}\, \scriptsize{\textbf{X-Men} (Q128452)\, |}\, \scriptsize{\textbf{Marvel Comics} (Q173496)\, |}\, \scriptsize{\textbf{To Be {[}music album{]}} (Q17025795)\, |}\, \scriptsize{\textbf{The Allman Brothers Band} (Q507327)\, |}\, \scriptsize{\textbf{proton–proton chain reaction} (Q223073)\, |}\, \scriptsize{\textbf{features of the Marvel Universe}(Q5439694)\, |}\, \scriptsize{\textbf{Features of the Marvel Cinematic Universe} (Q107088537)\, |}\, \scriptsize{\textbf{Uncanny X-Men} (Q1399747)\, |}\, \scriptsize{\textbf{member of parliament} (Q486839)}} \\
        \midrule
        TEC $E_{LM} \left(\alpha=0\right)$     & \multicolumn{1}{p{12cm}}{\scriptsize{\textbf{Google} (Q95)\, |}\, \scriptsize{\textbf{Amazon} (Q3884)\, |}\, \scriptsize{\textbf{Microsoft} (Q2283)\, |}\, \scriptsize{\textbf{open source} (Q39162)\, |}\, \scriptsize{\textbf{Apple Inc.} (Q312)\, |}\, \scriptsize{\textbf{Facebook} (Q355)\, |}\, \scriptsize{\textbf{Meta Platforms} (Q380)\, |}\, \scriptsize{\textbf{Cisco Systems} (Q173395)\, |}\, \scriptsize{\textbf{Salesforce.com} (Q941127)\, |}\, \scriptsize{\textbf{Citrix Systems} (Q916196)}}                                                                                                                                \\
        TEC $E_{G} \left(\alpha=\infty\right)$ & \multicolumn{1}{p{12cm}}{\scriptsize{\textbf{Mike Tyson} (Q79031)\, |}\, \scriptsize{\textbf{World Boxing Organization} (Q830940)\, |}\, \scriptsize{\textbf{International Boxing Federation} (Q742944)\, |}\, \scriptsize{\textbf{Floyd Mayweather} (Q318204)\, |}\, \scriptsize{\textbf{World Boxing Association} (Q725676)\, |}\, \scriptsize{\textbf{Tyson Fury} (Q1000592)\, |}\, \scriptsize{\textbf{Manny Pacquiao} (Q486359)\, |}\, \scriptsize{\textbf{World Boxing Council} (Q724450)\, |}\, \scriptsize{\textbf{Evander Holyfield} (Q313451)\, |}\, \scriptsize{\textbf{Joe Frazier} (Q102301)}}                                     \\
        \bottomrule
    \end{tabular}
    \caption{Example topics using WIKIPEDIA corpus for models trained with 300 topics. Each topic is represented by its top 10 entities. Here, the Q number represents the unique wikidata identifier.}
    \label{tab:sample_topics}
\end{table*}

\subsection{Quantitative results}

Table \ref{tab:coherence} shows that both LDA and the neural topic models underperform compared to TEC across all datasets and metrics, except for topic diversity of LDA. Entity extraction leads to sparser representations of corpora,  consequently resulting in poorer coherence than typically expected from topic models trained on word tokens --- a challenge that underscores the difficulty of constructing practical entity-based topic models. However, our proposed method significantly outperforms similar state-of-the-art approaches. While LDA is quite consistent in terms of topic diversity, the overall topic quality is generally poor, as evidenced by its coherence.
Our results suggest that the most valuable source of information for building entity-based topic models is explicit knowledge from a graph neural network as opposed to implicit knowledge from the language model. These findings are supported by the fact that TEC performs consistently worse when purely built using language model embeddings ($\alpha=0$) in comparison to models constructed exclusively with graph embeddings ($\alpha=\infty$) or some combination of the two ($\alpha \in \{1/2, 1, 2 \}$). Furthermore, as the scale of the contribution moves toward graph embeddings (represented by increasing values of $\alpha$), the overall performance rises steadily, supporting the conclusion that large language model embeddings are generally less important in entity-driven settings. We hypothesise that the increase in performance seen when using explicit knowledge representations can be attributed to their ability to capture semantically related concepts, providing supplementary information to these sparse entity-document terms. Concretely, the graph network is trained on concepts and those within its proximity in the knowledge base, which affords further context to these entity representations when performing clustering which results in better topics. However, we note that in some cases, implicit knowledge from a language model is not always detrimental as in the cases of \emph{CC-News} with $300$ topics or \emph{MLSUM} with $100$ topics. As such, its not completely inconceivable to image cases where such knowledge would be useful given certain metrics or corpora.

\subsection{Qualitative results}
We present exemplar topics for the different models in Table \ref{tab:sample_topics}. Here, we map these entities to their English form although we note that these concepts are language agnostic. Using visual inspection, we find cases where some top entities do not match the general topic theme. These must be attributed to limitations in the model as they all share the same preprocessed corpora, with the exception of WikiPDA. Overall these issues seem less prevalent with TEC. Particularly for ProdLDA and CombinedTM, we also find unrelated entities that linger across many topics, with the lingering entities varying between runs. We also find topics covering multiple themes, such as the ones resulting from LDA and WikiPDA.
\section{Limitations}
TEC assumes that documents contain entities, yet this is not necessarily the case. The proposed model is specifically valuable for entity-rich applications such as news articles.
A potential solution we are interested in exploring in the future is to train a self-supervised model to generate word embeddings using the bag-of-words as input and the document embedding as the target, to have words represented in the shared embedding space. 

We demonstrate that graph neural networks trained on a knowledge base of concepts are highly effective for constructing entity-based topic models. Whilst \emph{node2vec} has proven both practical and sufficient for our work, it is still a shallow neural network that may be unable to learn deeper, more complex relationships between entities. We believe that our results can improve if we obtain embeddings using a multilayer graph neural network with unsupervised training, although we note that the goal of this work is to explore entities as features for neural topic models, for which we demonstrate that graph embeddings trained on an explicit knowledge base performs best. 

Another issue, pervading topic modeling literature, is that current automatic coherency metrics may be poorly aligned with human judgment on topic quality \citep{hoyle2021broken, doogan2021twaddle}. Furthermore, these automated approaches rely heavily on co-occurrence statistics, suboptimal when working with sparse term-document frequencies. It would be interesting to evaluate these models directly using human annotators to judge the performance, for example using intrusion tests. 

Lastly, updating the knowledge base will force the retraining of the model, which does not currently guarantee a direct relationship between former and new topics. It requires additional research as this can be a hindrance for some applications.
\section{Conclusions}

We explore entity-based neural topic models based on the clustering of vector representations. Despite the sparse data structures which result from extracting entities from text data, we find that our TEC model, which performs clustering of implicit and explicit knowledge representations can produce more coherent topics than current state-of-the-art models. 
TEC represents documents using language-agnostic entity identifiers, resulting in a single set of topics shared across languages. It allows the extension to new languages without sacrificing the performance of the existing ones.

Our results suggest that the implicit knowledge provided by language models is superior to the state-of-the-art in terms of coherence and quality. Nevertheless, these results are surpassed by the explicit knowledge encoded in graph-based embeddings, using human contributed Wikidata knowledge base as a source.

\section*{Acknowledgements}

We thank Tiziano Piccardi, Federico Bianchi, and our colleagues at Huawei Ireland Research Centre, for helpful comments and discussion.

\section*{Bibliographical references}

\vfill
\pagebreak
\appendix
\FloatBarrier
\begin{table*}[!ht]
	\small
	\centering
	\begin{tabular}{l|rrr}
		\toprule
		\multicolumn{4}{c}{$\mathbf{CCNews}$}                                                                                   \\
		\toprule
		\textbf{Model}                         & $\mathbf{TC}$            & $\mathbf{TD}$            & $\mathbf{TQ}$            \\  \midrule \specialrule{.1em}{.05em}{.05em}
		\multicolumn{4}{l}{\textbf{Number of Topics} $\mathbf{\times 100}$}                                                     \\
		\midrule
		LDA (Entities)                         & $-0.13$ $(0.01)$         & $\textbf{0.97}$ $(0.00)$ & $-0.13$ $(0.01)$         \\
		LDA (Words)                            & $0.14$ $(0.00)$          & $0.59$ $(0.00)$          & $0.08$ $(0.00)$          \\
		\midrule
		TEC $E_{G} \left(\alpha=\infty\right)$ & $\textbf{0.20}$ $(0.02)$ & $0.83$ $(0.01)$          & $\textbf{0.16}$ $(0.02)$ \\
		\midrule \specialrule{.1em}{.05em}{.05em}
		\multicolumn{4}{l}{\textbf{Number of Topics} $\mathbf{\times 300}$}                                                     \\
		\midrule
		LDA (Entities)                         & $-0.07$ $(0.01)$         & $\textbf{0.91}$ $(0.00)$ & $-0.07$ $(0.01)$         \\
		LDA (Words)                            & $0.14$ $(0.00)$          & $0.43$ $(0.00)$          & $0.06$ $(0.00)$          \\
		\midrule
		TEC $E_{G} \left(\alpha=\infty\right)$ & $\textbf{0.18}$ $(0.01)$ & $0.76$ $(0.01)$          & $\textbf{0.14}$ $(0.01)$ \\\bottomrule
	\end{tabular}
	\caption{Results on CCNews corpus for all topic models. We record the results on  $\mathbf{TC:}$ (Topic Coherence), based on the normalized pointwise mutual information of top $N=10$ entities in a topic, $\mathbf{TD}:$ (Topic Diversity) the ratio of unique entities to total entities and $\mathbf{TQ:}$ (Topic Quality) Topic Diversity $\times$ $\mathbf{TC}$. The results are reported as the averages of 10 experimental runs and include the margin of error related to 95\% confidence intervals. Our model outperforms all baselines across all metrics except for $\mathbf{TD}$. \label{tab:ccnew_words}}
\end{table*}

\begin{table*}[ht]
	\centering
	\begin{tabular}{ccccccccccc}
		\toprule
		\multicolumn{11}{c}{$\mathbf{LDA (Word-Tokens)}$}                         \\
		\midrule \specialrule{.1em}{.05em}{.05em}
		Topic 1 &  & Topic 2  &  & Topic 3 &  & Topic 4 &  & Topic 5 &  & Topic 6 \\
		\midrule
		polic   &  & fire     &  & peopl   &  & counti  &  & appl    &  & price   \\
		offic   &  & weather  &  & thing   &  & st      &  & app     &  & averag  \\
		man     &  & storm    &  & make    &  & hill    &  & devic   &  & target  \\
		arrest  &  & hurrican &  & realli  &  & west    &  & phone   &  & analyst \\
		kill    &  & area     &  & talk    &  & ohio    &  & googl   &  & stock   \\
		\bottomrule
	\end{tabular}
	\caption{Example of the top 5 terms from six randomly sampled topics from LDA trained on word-level tokens. Given that this method is a common approach to topic modeling, it is unsurprising that the topics are quite coherent.\label{tab:ex_lda_words}}
\end{table*}

\begin{table*}[ht]
	\centering
	\begin{tabular}{ccccc}
		\toprule
		\multicolumn{5}{c}{$\mathbf{LDA (Entity-Tokens)}$}                                                    \\
		\midrule \specialrule{.1em}{.05em}{.05em}
		Topic 1             &  & Topic 2                     &  & Topic 3                                     \\
		\midrule
		Puerto Rico FC      &  & Las Vegas                   &  & WWE                                         \\
		Audit               &  & Nevada                      &  & Sunderland A.F.C                            \\
		broadcasting        &  & Hawaii                      &  & cycling                                     \\
		Victoria            &  & \makecell{University of                                                      \\ the Pacific} & & Orix Buffaloes \\
		farming             &  & Pixel 1 XL                  &  & Brock Lesnar                                \\
		\midrule
		Topic 4             &  & Topic 5                     &  & Topic 6                                     \\
		\midrule
		Aston Villa F.C     &  & Russia                      &  & White House                                 \\
		Birmingham City F.C &  & \makecell{Ultimate Fighting                                                  \\ Championship} & & presidency of Donald Trump \\
		Burton Albion F.C   &  & Vladimir Putin              &  & Minnesota                                   \\
		music interpreter   &  & Moscow Kremlin              &  & Melania Trump                               \\
		Midland             &  & mixed martial arts          &  & \makecell{Executive Office of the President \\ of the United States} \\
		\bottomrule
	\end{tabular}
	\caption{Example of the top 5 terms from six randomly sampled topics from LDA trained on entity tokens. Here we see that topics are not as clear when compared to a model trained on word-level tokens. \label{tab:ex_lda_entities}}
\end{table*}

\begin{table*}[!ht]
	\centering
	\begin{tabular}{ccccc}
		\toprule
		\multicolumn{5}{c}{$\mathbf{TEC \alpha=2}$}                                   \\
		\midrule \specialrule{.1em}{.05em}{.05em}
		Topic 1         &  & Topic 2                         &  & Topic 3             \\
		\midrule
		Brazil          &  & Bill Clinton                    &  & France              \\
		Rio de Janeiro  &  & Barack Obama                    &  & Paris               \\
		São Paulo       &  & Joe Biden                       &  & Claude Lelouch      \\
		Portugal        &  & presidential election           &  & Laurent Ruquier     \\
		Tony Ramos      &  & George W. Bush                  &  & Patrice Leconte     \\
		\midrule
		Topic 4         &  & Topic 5                         &  & Topic 6             \\
		\midrule
		Bill Gates      &  & Basketball                      &  & Iran                \\
		Warren Buffett  &  & National Basketball Association &  & Tehran              \\
		Mark Zuckerberg &  & LeBron James                    &  & Ali Khamenei        \\
		Steve Jobs      &  & Jason Kidd                      &  & Mahmoud Ahmadinejad \\
		Google          &  & Kevin Durant                    &  & Ruhollah Khomeini   \\
		\bottomrule
	\end{tabular}
	\caption{Example of the top 5 terms from six randomly sampled topics from TEC $\alpha=2$ trained on entity tokens. We see that our novel approach produces highly interpretable and coherent topics from entities. \label{tab:ex_tec}}
\end{table*}

\section{Entity extraction}
\label{app:entity-extraction}
\textbf{Pattern matching}. We first extract candidate entities by finding language-specific text patterns in the original text. For a given language, these text patterns are collections of preprocessed surface forms representing the multiple entities contained in a knowledge base. Inspired by \citet{mendes2011dbpedia} and \citet{daiber2013improving}, we use the deterministic Aho-Corasick algorithm \citep{aho1975efficient} due to its speed and effectiveness in extracting text patterns. The only language-specific components are the preprocessing components, such as lemmatizers, that increase the number of relevant entity matches. These preprocessing components are independent of each other. Consequently, we can expand the model to additional languages without compromising the performance of the others.
Assuming the KB contains surface forms in multiple languages, we build finite-state automata, one per language, using lists of preprocessed text patterns. The only language-specific components are the preprocessing components, such as lemmatizers, that increases the number of useful entity matches. This approach provides independence between languages which allows their expansion or update without risking the performance of the incumbent remaining ones.

An alternative method to extract entities is \textit{named entity recognition} (NER) and \textit{entity linking} (EL), but that requires more complex language-specific pipelines with unknown guarantees of success.
Similarly to BOW-based methods that have a fixed vocabulary, our approach also assumes a fixed number of entities contained in the KB.
For that reason, Aho-Corasick automata are deterministic solutions guaranteed to contain the same entities that form the KB.

Extracting entities does also not guarantee that entities are linked to the ones existing in the KB. It would also require us to either use pretrained NER models or \textit{entity linking} (EL) is an alternative method to obtain entities but, that typically first finds entities and only then attempts to link them to a KB, we have as a requirement that the entities must be represented in the knowledge base. This requirement results from the definition of topic models where the vocabulary must be defined \textit{a priori}.

\textbf{Disambiguation}.
As each text pattern can be associated with multiple entities it is necessary to identify which one is the best match and whether that entity is not a spurious match, but is significant to represent the document. The text embeddings result from applying a language model and the entity embeddings are assumed to be previously computed as described in \ref{sec:entity_representation}. We use cosine similarity to rank the prospective entities and keep the most similar one as long as its score is above some predefined threshold. Notice that if the Language Model supports multiple languages then the entities only require a single embedding to represent it. In practice we can have as many entity embeddings as the languages we want to cover, and so will the entitizer as long as there are language-specific Aho-Corasick automata.

We then select the ones that are above a predefined threshold value. Regarding the entities obtained through pattern matching, the same text pattern may relate to multiple entities. For example, it is not uncommon to find organizations sharing the same acronyms or people with the same name. When a text pattern relates to multiple candidate entities, this method also performs disambiguation as we only keep, at most, the most similar entity\footnote{While this approach may not be the most optimal way of providing entities with both high precision and recall, it does generate a sufficient number for performing our research.}.

\section{Words vs Entities}
\label{app:words_ents}
We have previously stated that in Section \ref{exp} it is not valuable to directly compare word-level topic modeling to entity-based topic modeling due to the sparsity issues associated with PPMI and due to the fact that these metrics have previously received criticism for their inability to accurately determine satisfactory interpretable topics. In the literature, classical topic modeling methods are generally dominated by word or phrase-level approaches in which certain preprocessing steps are used to filter and tokenize sentences into lists of document terms. Common techniques for filtering include approaches that attempt to remove noisy or unhelpful terms from the documents via techniques such as the removal of punctuation, purely syntactically functional terms (for example, stop words) or low-frequency terms. However, in comparison to entity-based extraction methods, these word-level document representations are much denser in terms of raw token count and shallower in terms of actual semantic information. For instance, entity extraction and tokenization methods may isolate an entity such as \textsc{New York City}, from which word-level approaches may similarly filter the terms "New", "York" and "City" (or just "New York" since the disambiguation process can determine this phrase to be the entity \textsc{New York City}) which individually are more fine-grained in terms of syntagmatic/paradigmatic information between documents whilst also being less rich in terms of pure semantic information. Indeed, the term "city" most likely appears many times across documents in a corpus whereas something more specific such as the phrase "New York" or "New York City" which gets classified as the entity \textsc{New York City} may only appear sparingly, further illustrating the problem with sparsity which any entity-based model (including the model presented in this work) must overcome, and for which models such as LDA have been shown to be less suited. Further issues emerge when directly measuring and comparing the intrinsic performance between word and entity-level-based approaches. Because \textit{Positive Pointwise Mutual Information} (PPMI) is generally employed as a proxy for measuring the coherence of a topic, the sparse structure of entity-level tokens will produce highly distinct distributions to that of word-token PPMI matrices. Consider the fact that n-grams in word tokens will have high joint probabilities, which boosts coherence metrics as described in Equations \ref{eq:coherence_per_topic} and \ref{eq:npmi}. However, these higher coherence scores do not necessarily reflect an increase in interpretability. Thus, while coherency metrics may offer a useful simulacrum for gauging interpretability and meaningful thematic structure, they are ultimately only useful when comparing different topic modeling methods within the same tokenization strategy. To further demonstrate this and for completeness, we examine how a model such as LDA compares when trained on words and entities to further illustrate our discussion.

\subsection{Preprocessing}
For our experiments, we compare only on \emph{CC-News} text data. Note that we can only perform these experiments on \emph{CC-News} or \emph{Wikipedia} since word-level topic modeling is not directly compatible with the multilingual setting of \emph{MLSUM}, a distinct advantage of using entities. We perform word tokenization before removing stop words, punctuation and tokens that represent digits. For stemming, we use the \emph{NLTK} implementation of the \emph{Porter Stemmer}\footnote{\url{https://www.nltk.org/_modules/nltk/stem/porter.html}}. Finally, we prune the tokens by i) removing tokens that occur less than 15 times in the corpus ii) removing tokens that occur in less than 20 documents, and iii) removing tokens that occur in greater than 50$\%$ of documents. Similarly, we use the \emph{Mallet} implementation of LDA \citep{mccallum2002mallet} to train the LDA model.

\subsection{Results}
Looking at Table \ref{tab:ccnew_words}, we see that, in comparison to LDA trained on entities, the word-level tokens provide stronger performance on a range of coherency metrics except on topic diversity, indicating its inability to thoroughly extrapolate thematic structure. We attribute this to multiple factors, including the fact that these metrics rely on pointwise mutual information from extremely sparse co-occurrence information, and that LDA itself relies heavily on these term-document distributions in text data. However, when comparing word-level word-based LDA to TEC (although as we state, we must be careful when making direct comparisons) we see that TEC outperforms LDA on topic coherence and topic quality, whilst retaining a high degree of topic diversity further illustrating its performance. With respect to entity-based models, the inability of LDA to build coherent topics is particularly striking, as opposed to TEC. We also see that the LDA model trained on words tends to have less variation across these metrics, highlighting the difficulty of training with sparse, information-rich document terms.

\subsection{Topic examples}
Next, we provide further examination through visual inspection. We note that the following analysis is purely visual in nature, although they are supported by the empirical findings in the paper. Observing the results in Table \ref{tab:ex_lda_words}, we see that LDA trained on word-level tokens produce fairly reasonably coherent topics from the data, as to be expected from the model. Comparing with results of LDA trained on entities as seen in Table  \ref{tab:ex_lda_entities}, we observe that the topics are quite unclear with a number of nosy terms that do not seem to relate well to the other terms since LDA is not entirely equipped to handle this type of data. However, looking at examples from Table \ref{tab:ex_tec} we see that all topics are highly interpretable and coherent, demonstrating the strength of our novel approach. In the future, crowdsourcing tasks such as word intrusion or topic-word matching from human annotators may provide better insight when evaluating these models \citep{lund2019eval, hoyle2021broken, doogan2021twaddle}.

\end{document}